# Topic Independent Identification of Agreement and Disagreement in Social Media Dialogue

**Amita Misra & Marilyn A. Walker**
Natural Language and Dialogue Systems Lab
Computer Science Department
University of California, Santa Cruz
maw|amitamisra@soe.ucsc.edu

## Abstract

Research on the structure of dialogue has been hampered for years because large dialogue corpora have not been available. This has impacted the dialogue research community's ability to develop better theories, as well as good off-the-shelf tools for dialogue processing. Happily, an increasing amount of information and opinion exchange occur in natural dialogue in online forums, where people share their opinions about a vast range of topics. In particular we are interested in rejection in dialogue, also called disagreement and denial, where the size of available dialogue corpora, for the first time, offers an opportunity to empirically test theoretical accounts of the expression and inference of rejection in dialogue. In this paper, we test whether topic-independent features motivated by theoretical predictions can be used to recognize rejection in online forums in a topic-independent way. Our results show that our theoretically motivated features achieve 66% accuracy, an improvement over a unigram baseline of an absolute 6%.

## 1 Introduction

Research on the structure of dialogue has been hampered for years because large dialogue corpora have not been publicly available. This has impacted the dialogue research community's ability to develop better theories, as well as good off-the-shelf tools for dialogue processing that account for the richness of human dialogue. Happily, an increasing amount of information and opinion exchange occurs in natural dialogue in online forums, where people can express their opinion on a vast range of topics from *Should there be more stringent gun laws?* to *Are school uniforms a good idea?* (Walker et al., 2012a). For example, consider the dialogic exchange in Fig. 1.

| Post **P**, Response **R** |
|---|
| **P1**: **Can the** government force abortion clinics to carry anti-abortion articles and papers**?** Or maybe force them provide a sonogram**?** Force them to have a psychologist on staff**?** Force them to have 3x3 foot posters of aborted babies on the wall**?** **Seems like** it makes more sense for a state to restrict something from the people rather than force the people to have something. No**?** |
| **R1**: **I don't see** why this matters. **Could you** please elaborate a little more, and in that elaboration, could you address why the government may require a private company to provide this commonly recommended medical remedy (plan b) **when it does not** do so with countless other common medically recommended remedies**?** |

Figure 1: Disagreement from 4forums.com. Possible features in **bold**.

In particular we are interested in the phenomenon of REJECTION in dialogue (Horn, 1989; Walker, 1996a), also called disagreement and denial. Our data show that the amount of disagreement in online ideological dialogues ranges from 80% to 90% across topic. Such data provides a rich resource for testing theoretical accounts of rejection, as well as for developing computational models of how to recognize rejection in dialogue. To date, rejection has received relatively little attention in computational models of discourse because of its rareness in task-oriented, tutorial or SwitchBoard style dialogue. Computational models of argumentative discourse do not typically attempt to account for rejection in dialogue, focusing instead on monologic sources displaying legal reasoning, logical accounts of rejection, or how to produce good arguments using natural language generation (Zukerman et al., 2000; Carenini and Moore, 2000; Wiley, 2005; Sadock, 1977).

Moreover, the theoretical literature strongly suggests that there should be topic-independent indicators of rejection. In work on politeness theory, rejection is a dispreferred response, predicting that rejection should be associated with markers of dispreferred responses such as disfluencies and hedging (Brown and Levinson, 1987). Work on negation specifies markers of negation and contrast such as *but* or *only* for different types of rejection, and work on discourse relations and their markers suggests that DENIAL is a type of COMPARISON relation (Horn, 1989; Groen et al., 2010; Webber and Prasad, 2008). These observations, among others, suggest a range of theoretically motivated features for the classification of rejection in online dialogue, e.g. phrases such as *I think*, *but*, *I don't see*, and *Can you*. See Fig. 1.

| Type | Context | Rejection |
|---|---|---|
| DENIAL | Pigs can fly. | **No**, you idiot, pigs **can't** fly! (Horn's 29) |
| LOGICAL CONTRADICTION | Kim and Lee have been partners since 1989. | But Lee said they met in 1990. |
| IMPLICIT DENIAL | Julia's daughter is a genius. | Julia **doesn't have** any children. |
| REFUSAL | Come and play ball with me. | **No**, I don't want to. (Horn's 33) |
| IMPLICATURE REJECTION | There's a man in the garage. | There's something in the garage. (Walker's 6) |
| DENYING BELIEF TRANSFER | B: Well ah *he uh ... he belongs to a money market fund now and uh they will do that for him.* H: The money market fund will invest it in government securities as part of their individual retirement account – is that what you're saying? B: Right. | H: **I'm not so sure of that.** (Walker's 31) |
| INCONSISTENT PAST BELIEF | H: Then they are remiss in not sending it to you because that money is taxable sir. | M: I know it's taxable, **but I thought** they would wait until the end of the 30 months. |
| CITING CONTRADICTORY AUTHORITY | H: No sir.... | R: That's what **they told me**. |

Figure 2: Classification and Examples of the Types of Rejections.

Our aim is to test whether theoretical predictions and topic-independent features motivated by them can be used to recognize rejection in online forums. We generalize our topic independent features using a development set on the topic *Evolution*. We then test a rejection (disagreement) classifier trained on *Evolution* on 1757 posts covering a collection of other topics, and compare our results to a ngram model trained on *Evolution* and tested on the same test set. See Table 1.

We first describe our corpus in Sec. 2, and then review previous work characterizing the theoretical basis of rejection in dialogue in Sec. 3. Sec. 4 describes our method for classifying rejections and Sec. 5 presents our results, showing that our theoretically motivated rejection cues are reliable across topic. We show that cue words, polarity, punctuation, denial and claim features motivated by the theoretical literature provide a significant improvement over a 50% baseline, and that all of the theoretically motivated features combined achieve 66% accuracy as compared to a unigram accuracy of 60%. We delay reviewing previous computational work rejection to Sec. 6 when we can compare it with our own work.

| Topic | Agr | DisAgr | Total |
|---|---|---|---|
| Evolution | 460 | 460 | 920 |
| Abortion | 250 | 280 | 530 |
| Climate Change | 17 | 10 | 27 |
| Communism vs. Capitalism | 10 | 13 | 23 |
| Death Penalty | 15 | 19 | 34 |
| Existence Of God | 53 | 48 | 101 |
| Gay Marriage | 173 | 134 | 307 |
| Gun Control | 334 | 331 | 665 |
| HealthCare | 21 | 37 | 58 |
| Marijuana Legalization | 6 | 6 | 12 |
| All Topics (test set) | 879 | 878 | 1757 |

Table 1: Distribution of (Dis)Agreement by Topic. The Evolution topic is for development and training. The test set of other topics is balanced overall, but not by topic.

## 2 Corpus

We utilize the publicly available Internet Argument Corpus (IAC), an annotated collection of 109,553 forum posts (11,216 discussion threads)(Walker et al., 2012a). We use the portion of the IAC containing dialogues from http://4forums.com. On 4forums, a person starts a discussion by posting a topic or a question in a particular category, such as society, politics, or religion. Forum participants can then post their opinions, choosing whether to respond directly to a previous post or to the top level topic (start a new thread). Conversants may simply agree or disagree with a previous post or they may provide a reasoned argument.

The corpus contains posts on topics such as *Abortion, Evolution, Existence of God, Gay Marriage* and *Gun control* along with a range of useful annotations. First, there are annotations that collapse different discussions into a single topic for 14 topics. For example, the *Evolution* and *Gun Control* topics include discussions initiated with the range of titles in Table 2, which guaran-tees variation in the focus of the discussion even within topic. The topics we use are in Table 1. Each discussion is threaded so that we can identify direct responses. Discussions may have a tree-like structure, so a post may have multiple direct responses. In addition to the adjacency pairs yielded by threading, `4forums` also provides a quote/response **Q/R** mechanism where a post may include a quote of part or all of a previous post. We do not use the Q/R pairs here.

| **First Post (P), Response (R)** |
|---|
| **Disagreements** |
| **P1:** No I didn't miss it, I was hoping you'd actually put forward an argument against what I said, not what you think I said. See what I actually said was the tautology. Then make your argument. Note Post 30 He said evolution is a tautology. I said that Darwin preferred a tautology to "Natural Selection" You may have mixed up who it is you're arguing against.<br>**R1: I'm wondering**. What do we call someone who debates feverishly on scientific theories, yet admittedly **does not** understand the concepts they are arguing against? Is it productive to debate something that **you don't understand** the concepts of when it's a fairly involved theory based on scientific evidence? What **if you** convinced someone NOT to believe in it, **but** you did so using falsifiable reasons, **since** you aren't an expert and might not know any better**?** Irresponsible, is one such word, that comes to mind. |
| **P2:** What in Vishnu's name does this have ANYTHING to do with evolution vs creation???<br>**R2: Well**, many have argued that **if you don't** except a literal Genesis, you're damned. **Perhaps** not in this particular thread, **but** the arguments are **essentially** the same. **I believe that** the theological implications of that position are fair game for discussing the validity of creationism. |
| **P4:** You have this backwards. The word theory was originally a scientific word, and then it was adapted into common speech to mean a range of things not originally designated to that word. Words like evolve, gravity and congruent have different meanings within the realm of science than they have outside. If you can't appreciate the difference between the definition of a word in the context of science as opposed to the context of common speech, then maybe you have no business in science.<br>**R4:** When it comes to all the examples that Behe had provided in both his first book, and his second book , **it has been shown** to be able to evolve naturally. That means, in principle, IC systems can evolve. **If you don't** believe so, bring forth the I.C. system of your choice. To say '**you don't know** all the answers' is just the logical fallacy known as 'argument from ignorance'. Behe brings a system up that he claims is IC. the pathway for evolution is discovered, and Behe trys another one. How dishonest **can you** get? The concept is falsified. |
| **P5**: Well, Genesis has God making all the animals "and their kind", and then when he's done with that he makes humans. So I would assume that humans don't fit into the "kind" schema, or perhaps are a kind unto themselves........<br>**R5:** : **So we can't** base our definition of "kind" on mere appearances? **I mean** if we are going to put things into categories and call the category "kind", **we should** do this by common appearances. A penguin is in the same kind as a hummingbird, **but** is a lobster in the same kind as an oyster? ........ |
| **Agreements** |
| **P6**: I think its nonsense interpretation developed by people who were afraid that if they fought for guns as valiantly as they did for free speech, they wouldn't receive any donations.<br>**R6**: I think you are entirely correct. From the page VOR linked: There is no evidence ANYWHERE that the second amendment is a collective right. We have been over this multiple times, and the evidence simply does not exist, and an organization like the ACLU should be well aware of this. |
| **P7:** Correction: If one isn't a fundementalist, literal christian, jew or muslim, then marc considers them a atheist. He's never going to deal with the fact that he's quite wrong on that subject. It's obvious to everyone that he's constantly avoiding it even when asked point blank several times. A sign of argumental failure is constant avoidance of a simple question.<br>**R7:** Quite **right**. My mistake. Once again, quite **right**... |
| **P8:** thats pretty neat. Did they finish up the feeder?<br>**R8: yeah**, this is clearly the best thread on these forums in probably the past year....give us some more pics length) |
| **P9**: This is probably the most rational site in all of the creationist's online arguments. Arguments we think creationists should NOT use<br>**R9: Thanks**, DuoMax, for this link. How delightful to see here mention of this solid gesture, on the part of a major creationist organization, in the direction of intellectual integrity..... ....Each time a Christian stands in the pulpit and pours out poor argument, s/he loses ground for the faith. **Thanks** again. |

Figure 3: Disagreements and Agreements from `4forums.com`. Theoretically motivated features are in **bold**.

| | |
|---|---|
| Evolution | Evolution in school, Dinosaurs and Human Footprints, Can Evolution & Religion Coexist, Did Charles Darwin Recant, Shrinking Sun, Bombardier beetle, Moon Dust, Second Law of Thermodynamics, Magnetic Field, Nebraska Man |
| Gun Control | Gun Control, Trigger Locks, Guns in the Home, Right to Carry, Assault Weapons, One gun a month, Gun Buy Back, Gun-Seizure Laws, Plastic Guns, Does gun ownership deter crime, Second Amendment, Enforced Gun Control Laws?, Gun Registration, Armor piercing bullets, Background Checks at Gun Shows |

Table 2: Discussions Mapped to the Evolution and Gun Control Topics.

The IAC also includes annotations collected via Mechanical Turk on these dialogue pairs. There are 20,000 pairs from threads of 3 posts P1,P2,P3 with annotations for (dis)agreement for pairs (P1, P2) and (P2, P3). Agreement was a scalar judg-ment on an 11 point scale [-5,5] implemented with a slider. The annotators were also able to signal uncertainty with a CAN'T TELL option. Each of the pairs was annotated by 5-7 annotators, in response to the annotation question *Does the respondent agree or disagree with the prior post?*. Annotators achieved high agreement on dis(agreement) annotation with an $\alpha$ of 0.62. We used thresholds of 1 and -1 on the mean agreement judgment to determine agreement and disagreement respectively. We omitted dialogue adjacency pairs with mean annotator judgment in the (-1,1) range. Table 1 provides the distribution of topics for the 1757 posts in the test set.

## 3 Theories of Rejection in Dialogue

A common view of dialogue is that the conversational record is part of the COMMON GROUND of the conversants. As conversants A and B participate in a dialogue, A and B communicate through dialogue speech acts such as PROPOSALS, ASSERTIONS, ACCEPTANCES and REJECTIONS. If A asserts a proposition $\phi$ and B accepts A's assertion, the $\phi$ becomes a mutual belief in the common ground. If B rejects A's assertion or proposal, the common ground remains as it was (Stalnaker, 1978). For conversants to remain coordinated (Thomason, 1990), they must monitor whether their utterances are accepted or rejected by their conversational partners.

Computational models of dialogue also must track what is in the common ground (Traum, 1994; Stent, 2002). This would be simple if conversants always explicitly indicated rejection with forms such as *I reject your assertion*. However recognizing rejection typically relies on making inferences. Horn categorizes rejections into: DENIAL a straightforward negation of the other's assertion; LOGICAL CONTRADICTION following from logical inference; IMPLICIT DENIAL where B denies a presupposition of A's; and REFUSAL, also called REJECTION where B refuses an offer or proposal of A's (Horn, 1989). See Fig. 2. All of Horn's forms can be identified as rejections by recognizing logical inconsistency either directly from what was said, or via an inferential chain.

However subsequent work by Walker on the *Harry Gross Corpus* (henceforth **HGC**) of advice-giving dialogues (Pollack et al., 1982) demonstrated that REJECTION IMPLICATURES as seen in the 5th row of Fig. 2, are common in natural dialogue (Walker, 1996a). A number of similar examples can also be found in (Hirschberg, 1985). Here, the proposition realized by the response follows from the original assertion as an entailment via existential generalization. Thus the REJECTION IMPLICATURE is logically consistent with the original assertion.

Walker argues that the fact that an implicature can function as a rejection clearly indicates that inference rules about what gets added to the common ground must have the same logical status as implicatures, i.e. they must be default rules of inference that can be defeated by context. She then goes on to identify additional types of rejections in **HGC** that rely on detecting conflicts in the default inferences triggered by the epistemic inference rules used in speech act theory. Walker uses a compressed version of rules from (Perrault, 1990; Appelt and Konolige, 1988), assuming that conflicting defaults can arise between these inferences and implicature inferences (Hirschberg, 1985). The first rule is given in 1:

(1) BELIEF TRANSFER RULE:
Say(A,B,p) $\rightarrow$ Bel (B,p)

The Belief Transfer Rule states that if one agent A makes an assertion that p then by default another agent B will come to believe that p. The second rule is in 2:

(2) BELIEF PERSISTENCE RULE:
Bel (B,p,$t_0$) $\rightarrow$ Bel (B,p,$t_1$)

The Belief Persistence Rule states that if an agent B believes p at time $t_0$ then by default agent B still believes p at a later time $t_1$. These rules provide the basis for inferring three additional types of rejections:

- DENYING BELIEF TRANSFER: Agent B can deny the consequent of the Belief Transfer Rule by negatively evaluating A's assertion or expressing doubt as to its truth.
- INCONSISTENT PAST BELIEF: Inferring that B's expression of an inconsistent past belief is a type of rejection relies on detecting conflicting defaults with the Belief Transfer Rule and the Belief Persistence Rule. The two beliefs may directly conflict, or the conflict may arise via an inferential chain.
- CITING CONTRADICTORY AUTHORITY: Inferring that citing a contradictory authority is a type of rejection relies on recognizing two inconsistent instantiations of the Belief Transfer rule. For example, agent A1 asserted p and agent A2 asserted $\neg$p, leaving B in an inconsistent belief state caused by the conflicting defaults generated by the alternate instantiations of the Belief Transfer Rule.

Fig. 2 provides Walker's examples of these new types of rejection and Fig. 3 illustrates disagreements and agreements in the IAC corpus.[1] While we see many instances of the rejection types in Fig. 2 in IAC, especially CITING CONTRADICTORY AUTHORITY and DENYING BELIEF TRANSFER, we also find new types such as ad-hominem attacks on the other speaker as the source of particular propositions (e.g. **R1** in Fig. 3, which would not have occurred in **HGC** talk show context. Other cases that we have noted are a different type of DENYING BELIEF TRANSFER, which occurs when a previous speaker's asserted proposition is marked by the hearer as hypothetical using a conditional, e.g. *If capital punishment is a deterrent, then .....* In future work we aim to expand the taxonomy of rejections using IAC.

## 4 Empirical Method

Our primary hypothesis is that certain expressions and phrases are reliable cues to the automatic identification of the speech acts of REJECTION and ACCEPTANCE, i.e. (dis)agreement, independently of the topic. We assume that it will not always be possible to get annotated data for a particular topic, given the ever-burgeoning range of topics discussed online. We use the *Evolution* topic as our development set, and ask: given (dis)agreement annotations for only one topic, is it possible to develop features that perform well on another arbitrary topic?

There is limited previous research on disagreement, thus it is an open issue what types of features might be useful. One line of previous work suggests that various pragmatic features might help (Galley et al., 2004). Another line suggests that disagreement is subtype of the COMPARISON (CONTRAST) discourse relation, in the Penn Discourse TreeBank taxonomy, suggesting that features for identifying COMPARISON, such as polarity and discourse cues might also be useful (Hahn et al., 2006; Prasad et al., 2010; Louis et al., 2010).

We began by selecting and manually inspecting 460 agreements and 460 disagreements from the *Evolution* topic, and extracting their most frequent unigrams, bigrams and trigrams. This showed that features suggested by theoretical work on rejection were indeed highly frequent: our aim was to generalize what we observed in the *Evolution* dataset and then test whether the generalized features can distinguish agreements from disagreements. We first observed that very few unigrams were useful for disagreements, e.g. *liar*, *no*, *don't*, while bigrams such as *I don't*, *How can*, *If I*, *how could*, *show me* seemed to be better indicators. Furthermore, trigrams such as *I don't agree*, *how can you*, *point is that*, and *I do not understand* are even stronger indicators of disagreement, but of course these higher order ngrams are less frequent and are more likely to contain topic-specific words. In order to provide better generalization, we generalized the ngrams that we observed, e.g. an instance such as *how can you* would also result in *how can we* and *how can they* being added to the same feature set. We also generalized over hedges and other categories of features on the basis of the theoretical literature. The total set of features we developed are grouped into the sets in Table 3 discussed in detail below.

| Feature | Description | Examples |
|---|---|---|
| Agreement | Ngrams indicative of accepting others claim. | *right*, *yes*, *yeah*, *correct*, *accepted*, *thanks*, *good*, *agree*, *acknowledge* |
| Cue Words | Cues as Ngrams and their LIWC CogMech generalizations | *oh*, *so*, *uh*, *yes*, *no*, *dont*, *cogmech*, *claim*, *i*, *yeah*, *because*, *well*, *just*, *and*, *you*, *you mean*, *i see*, *i COGMECH* |
| Denial | Ngrams indicative of denying another's claim | *You don't know*, *That does not*, *I don't think*, *what is*, *This has nothing*, *I don't see*, *You do not*, *do you mean*, *I don't know*, *we don't have*, *Problem with that*, *I do not*, *Does not*, *why do*, *But I don't*, *how can* |
| Hedges | Unigrams, bigrams, and trigrams that include hedge terms. | *Im wondering*, *I am wondering*, *whatever*, *somewhat*, *may be*, *possibly*, *anyway*, *it seems to me*, *my view*, *actually*, *my opinion*, *essentially*, *somewhat*, *my perspective*, *rather*, *although*, *really*, *I suppose*, *perhaps* |
| Duration | Sentence, word and post lengths | |
| Polarity | Means of positive and negative polarity terms. | |
| Punctuation | Counts of question marks and exclamation points. | |

Table 3: Feature Sets, Descriptions, and Examples. The unigrams features are our baseline case; these features are not theoretically motivated.

**Unigrams.** Results of previous work on stance identification in argumentative discourse suggest that a unigram baseline can be difficult to beat (Thomas et al., 2006; Somasundaran and Wiebe, 2010). Thus we test our theoretically motivated features against unfiltered unigrams and un-

[1] Since participants are not generally making plans together in these dialogues, we leave aside Walker's classification of rejections of proposals.

igrams+bigrams as baselines.

**Agreement and Denial.** As described above we used *Evolution* to manually develop generalizations of the observed unigrams, bigrams and trigrams that were consistent with theoretical predictions. We split the indicator features into two categories Agreement and Denial. See Table 3. Our manual analysis suggested that agreements have few topic independent markers. Unigrams such as *agree correct* and *right* were also present in disagreements, and trigrams such as *I agree but*, *You may be correct however I do not agree*, *I don't agree* were better indicators of disagreement. Our agreement markers are thus a small category where we check that the keywords *agree, correct* and *right* are not preceded by a negation marker and not followed by discourse markers such as *but*, *yet*, or *however*. However, the denial category at present has more than 300 ngrams extracted and generalized from the *Evolution* topic. Pitler et al, (2009) also used ngrams consisting of the first and last three words for recognition of the PDTB COMPARISON relation. Other work on the PDTB also suggests that DENIAL can be indicated by contrast (Webber and Prasad, 2008).

**Cue Words.** Both psychological research on discourse processes (Fox Tree and Schrock, 1999; Groen et al., 2010) and computational work on agreement and discourse markers (Galley et al., 2004; Louis et al., 2010) indicate that discourse markers are strongly associated with particular pragmatic functions such as stating a personal opinion (Asher et al., 2008; Webber and Prasad, 2008). Based on manual inspection of the *Evolution* devset we selected 18 items for the CUE WORDS feature set, as in Table 3. Examples are *well* in **R2** and *so* and *but* in **R5**.

**Durational Features.** Brown and Levinson's theory of politeness would suggest that disagreements are dispreferred responses and thus that the length of the post could indicate disagreement; it predicts that people will elaborate more and provide reasons and justifications for disagreement (Brown and Levinson, 1987). Our durational features measure the length of the utterance in terms of characters, words and sentences.

**Hedges.** In Brown and Levinson's theory of politeness, hedges are one of many possible strategies for mitigating a face-threatening act (Brown and Levinson, 1987; Lakoff, 1973). Hedges can be used to be deliberately vague or simply to soften a claim. We see many examples of hedges in online dialogue, e.g. the speaker of **R2** in Fig. 3 uses the hedges *Perhaps* and *essentially*, and *I mean* in **R5**. Thus hedges are hypothesized to be useful feature for distinguishing (dis)agreement, yielding the hedge features in Table 3.

**Polarity.** Work on discourse relations in the PDTB also suggests that differences in polarity across adjacent utterances might be an indicator of the COMPARISON relation. In addition, Horn's classes of REJECTIONS shown in Fig. 2 all include markers of negation. Thus to capture the overall sentiment of the post we used the MPQA subjectivity lexicon (Wiebe et al., 2003; Wilson et al., 2005). Each word is POS tagged and then categorized as strongly or weakly subjective. The positive polarity feature is the sum of the strongly subjective words of positive polarity, and the negative polarity feature represents the sum of strongly subjective words of negative polarity.

**Punctuation.** Another indication of DENYING BELIEF TRANSFER rejections are the question marks and exclamation marks that conversants frequently use to express their disbelief and doubt about another conversant's claim. For example, **R1** and **R5** in Fig. 3 have a high frequency of question marks.

## 5 Results

Our aim was to test how well we can distinguish agreements and disagreements in IAC using classifiers trained with theoretically motivated features. As described above, we developed our features by manual inspection of (dis)agreements in 920 posts on the topic *Evolution*. We do not train on a mixture of topics for any feature set, including unigrams, because we assume that in general, new topics are always arising so there will not be annotated data for every topic. We evaluate the performance of all types of features on classifying (dis)agreements on other topics combined. We do not report per-topic results because our test set baseline accuracies vary a great deal by topic as do the size of the topic sets. See Table 1.

| Features | Random Forest | J 48 |
|---|---|---|
| **ALL-TM** | 63.1 | 66.0 |
| **Unigram** | 56.6 | 59.8 |
| **Bigram** | 59.3 | 60.1 |

Table 4: Accuracies for Theoretically Motivated Features (**ALL-TM**), Unigrams and Bigrams with Random Forest and J48 Trees over a 50% baseline. No interesting differences observed in precision and recall.

Table 3 summarizes our theoretically-motivaed topic-independent features, and Table 4 compares the accuracies of classifiers using these features to unigrams and bigrams when we train on *Evolu-* *tion* and then test on our mixed-topic test set, using the Weka learners for Random forest and J48 Tree. Although unigrams and unigram+bigram achieves approximately 60% accuracy over a 50% baseline, paired t-tests on the result vectors show that the differences in accuracies are statistically significant when we compare ALL-TM features with unigrams and unigram+bigrams: Random Forest (p = .004) and J48 Trees ($p < .0001$).

| Ngram | N Feats | Acc | Feats Selected |
|---|---|---|---|
| **Uni** | 2K | 62.5 | *understand*, *fail*, *never*, *nothing*, **catholic**, **gene**, *irrelevant*, *acceptable*, *show*, *didn't*, **geologist**, **creationist** |
| **Bigram** | 4k | 62.7 | *? you*, *do we*, *understand that*, *? just*, *really?*, *is based*, *well said*, *? did*, *can the*, *the nature*, **the church**, *failed to*, *then what* |

Table 5: Accuracy when fitting to test set for number of features selected for ngrams, with sample features.

Moreover even if we optimize on the test set by examining the variations in performance as a function of the number of features selected, ALL-TM still beats both unigram and unigram+bigram, when features are selected according to ranking by Gain Ratio. ALL-TM is significantly more accurate when compared to unigrams (p = .003) best accuracy of 62.5 with 2000 features, and better than unigram+bigram best accuracy of 62.7 for 4000 features (p = .007). See Table 5.

More interestingly though, if we look at what features get selected ( Table 5), we see many features reminiscent of our theoretically motivated features. Features highly ranked by the Gain Ratio were topic-independent cues for disagreement such as *understand*, *fail*, *nothing*, *never* and Bigrams such as *? how*, *perhaps you*, *would you*, *never said*. However there were few high ranked unigrams and bigrams for agreement. Also note that topic specific cues such as *gene*, *catholic*, *creationist*, *geologist* and *the church* are selected over any topic-independent cues for agreement. This corroborates our manual construction of a combined denial category with more than 300 words and a very limited agreement category.

To test which features make the most difference, we also conducted ablation experiments (Table 6), as well as tests with individual features (Table 7). Table 6 shows that the CUE WORDS (p = .0008) and PUNCTUATION features (p = .01) have the biggest impact on performance. The decrease in performance when ablating agreement features is not statistically significant (p = .20).

| Ablated Feature | Random Forest | J 48 |
|---|---|---|
| **No Agreement** | 62.2 | 65.0 |
| **No Cue Words** | 59.1 | 62.1 |
| **No Denial** | 63.3 | 66.0 |
| **No Duration** | 63.6 | 66.3 |
| **No Hedges** | 64.2 | 66.5 |
| **No Polarity** | 64.4 | 66.8 |
| **No Punctuation** | 60.3 | 61.6 |

Table 6: Accuracy when Ablating each Theoretically Motivated Feature with Random Forest and J48 Trees over a 50% baseline .

| Feature | Acc | Prec | Recall |
|---|---|---|---|
| **Agreement** | 54.4 | .55 | .54 |
| **Cue words** | 62.5 | .63 | .62 |
| **Denial** | 52.0 | .54 | .52 |
| **Duration** | 53.6 | .54 | .53 |
| **Hedges** | 50.4 | .51 | .50 |
| **Polarity** | 53.4 | .53 | .53 |
| **Punctuation** | 65.3 | .65 | .65 |

Table 7: Results for Individual Features for J48 Trees over a 50% baseline .

Since the J48 learner performs consistently better, we restrict our comparison of individual features in Table 7 to that learner. Table 7 shows that PUNCTUATION and CUE WORDS features by themselves provide significant performance improvements over the unigram baseline, and that the POLARITY, AGREEMENT, DENIAL and DURATION feature sets provide significant improvements on their own over the majority class baseline of 50%. A paired t-test shows these differences are significant at p =.02. To our surprise, the HEDGE feature was not effective, and we plan further refinements of it. These results support the hypothesis that there are clearly markers for agreement and disagreement that are suggested by the theoretical literature and which are not topic specific.

## 6 Discussion and Future Work

We develop topic-independent features for classifying (dis)agreement in online dialogue, and show that we can beat an unfiltered unigram baseline by 6%, and even beat the best feature-selection ngram-based classifers fitted to the test set.

Features we didn't use from previous work include word pairs as introduced by (Marcu and Echihabi, 2002), and used subsequently by (Pitler et al., 2009) and (Biran and Rambow, 2011). The issue of whether word pairs are topic-dependent has never been addressed, but the examples given in previous work suggest that they may indicate topic-specific comparisons. Previous work also suggests that context might be helpful in recognizing disagreement (Walker et al., 2012b), but we did not test the effect of context.

The most similar work to our own trains a disagreement classifier for **Q/R** response pairs in online forums (Abbott et al., 2011). Their work used ngrams, MPQA opinion words (Stoyanov et al., 2005), LIWC (Pennebaker et al., 2001), and a different dataset (**Q/R** instead of **P1,P2** datasets), and does not aim to develop a classifier that works well independently of topic. Their best accuracy is about 68% for a feature set called BothLocal for the JRip classifier using $\chi^2$ feature selection. BothLocal includes unigrams, bigrams, trigrams, LIWC, punctuation, cue words, dependency features, generalized dependency features and utterance length measures, and it is unclear whether these features are specific to topic. It is also difficult to directly compare the results because they do not report accuracies for individual feature sets or ablated feature experiments. For example, their unigram accuracy of 63% includes cue words, and is reported for training and testing on a mixture of topics without any held-out topics.

Other work on disagreement recognition includes that of (Wang et al., 2011) who describe conditional random field model for detecting (dis)agreement between speakers in English broadcast conversations. They use sampling and prosodic features such as pause, duration and speech rate on an unbalanced dataset. They report an increase in F-measure of 4.5% for agreement and 4.7% for disagreement over a baseline of lexical, structural, and durational features. (Hahn et al., 2006) show that a contrast classifier improves the accuracy of dis(agreement) classification in the ICSI meetings corpus, and that their results are less affected by imbalanced data. They improve the F-measure to .755 over a baseline SVM with F-measure .726. (Yin et al., 2012) use sentiment, emotion and durational features for (dis)agreement classification in online forums, and they show that aggregating local positions over posts yields 3 to 4% better performance than non-aggregating baselines.

While recognizing (dis)agreement can be useful in its own right, it has also been shown to be useful for the identification of stance (Gawron et al., 2012; Hassan et al., 2010; Thomas et al., 2006; Bansal et al., 2008; Murakami and Raymond, 2010; Agrawal et al., 2003). Work that focuses on the social network structure of online forums as a way to improve stance classification has either assumed that adjacent posts always disagree, or used simple rules for identifying agreement based on patterns in the reply post (Murakami and Raymond, 2010; Agrawal et al., 2003). Previous work by Somasundaran & Wiebe (2009, 2010) develops positive and negative arguing features for the classification of stance, that at least in motivation, resemble our denial features . They show that arguing features are helpful in stance classification. Work by (Galley et al., 2004) on detecting disagreement in meetings corpora similarly shows that pragmatic features are useful for detecting disagreement using models based on Bayesian Networks. (Walker et al., 2012b) use a number of linguistic features such as unigrams, bigrams, and repeated punctuation and proposed a supervised model for stance classification in online debates. Related work by (Hassan et al., 2010) focuses on identifying the attitude of the participants towards one another in online debates. They relate the polarity of words to the second person pronoun for classification, while related work by (Abu-Jbara et al., 2012) uses the polarity of expressions and named entity recognition to identify a subgroup of participants, where participants within a subgroup are inclined to agree with one another. Methods for stance classification in congressional debates do not separately evaluate the accuracy of (dis)agreement classification (Thomas et al., 2006; Bansal et al., 2008; Awadallah et al., 2010; Burfoot, 2008).

In future work, we plan to develop more detailed patterns based on LIWC categories and syntactic parses (Thelen and Riloff, 2002). For example, an error analysis suggests that sometimes two people mutually reject the proposal or claim of a third person, e.g. *How can* **they** *say that....* In such cases our classifier finds the disagreement marker *how can* and classifies it as disagreement. More detailed syntactic processing would allow us to refine our patterns to identify particular classes of targets such as third person vs. first person. Similarly, here we extended patterns by hand, e.g. generalizations over pronouns such as *I can't, we can't, can you, can we*. In future we aim to generalize such patterns automatically using tagsets. We expect that more general patterns should improve the accuracy of the topic-independent feature sets. We also plan to carry out further annotation of the IAC corpus using the classes of rejections summarized in Fig. 2 to determine whether there are forms for indicating each type that are not represented by our features, and to determine the frequency across a sample of our corpus of the different types.